\documentclass[10pt,twocolumn,letterpaper]{article}

\usepackage{iccv}
\usepackage{times}
\usepackage{epsfig}
\usepackage{graphicx}
\usepackage{amsmath}
\usepackage{amssymb}
\usepackage{float}
\usepackage{algorithm}
\usepackage{tabularx}
\usepackage{algorithmic}
\usepackage{caption}
\usepackage{multirow}
\usepackage{algorithmic}
\usepackage{subfigure}
\usepackage{subfiles}
\usepackage{enumerate}
\usepackage{enumitem}
\usepackage{etoolbox,lipsum}
\usepackage[font=small,labelfont=bf,tableposition=top]{caption}
\usepackage{mwe}
\usepackage{balance}
\usepackage[titletoc,title]{appendix}
\pdfoutput=1

\newcommand{\SKIP}[1]{} % Used to skip stuff we do not want type-set

\newcommand{\mbegin} {\left [ \begin{array}}

\newcommand{\mend}   {\end{array} \right ]}

\newcommand{\detbegin} {\left | \begin{array}}
\newcommand{\detend}   {\end{array} \right |}

\newcommand{\vbegin} {\left ( \begin{array}{c}}

\newcommand{\vend} {\end{array}\right )}

\def\squareforqed{\hbox{\rlap{$\sqcap$}$\sqcup$}}

\def\qed{\ifmmode\squareforqed\else{\unskip\nobreak\hfil
	\penalty50\hskip1em\null\nobreak\hfil\squareforqed
	\parfillskip=0pt\finalhyphendemerits=0\endgraf}\fi}

\newcommand{\showeqnlabel}{
	\hbox to 0pt{\quad\quad\relax\fbox{\scriptsize\rm\eqnlblx}%
	\gdef\eqnlblx{xxxx}}} \newcommand{\eqnlblx}{}

\def\@eqnnum{\rm (\theequation)\showeqnlabel}

\newcommand{\nofig}[1]{\centerline{\bf Figure here}}

\def\mat#1{\mathchoice{\mbox{\bf$\displaystyle\tt#1$}}
	{\mbox{\bf$\textstyle\tt#1$}}
	{\mbox{\bf$\scriptstyle\tt#1$}}
	{\mbox{\bf$\scriptscriptstyle\tt#1$}}}
\def\m#1{\protect\mat #1}

\usepackage[pagebackref=true,breaklinks=true,letterpaper=true,colorlinks,bookmarks=false]{hyperref}

\iccvfinalcopy % *** Uncomment this line for the final submission

\def\iccvPaperID{2343} % *** Enter the ICCV Paper ID here

\ificcvfinal\fi
\begin{document}

%%%%%%%%% TITLE
\title{Monocular Dense 3D Reconstruction of a Complex Dynamic Scene\\from Two Perspective Frames}
\author{Suryansh Kumar$^1$
\and
Yuchao Dai$^{1,2}$
\and
Hongdong Li$^{1,3}$\\
\and
{$^1$Australian National University, Canberra, Australia} \\
{$^2$Northwestern Polytechnical University, Xi'an, China} \\
{$^3$Australia Centre for Robotic Vision} 
}

\twocolumn[{%
\renewcommand\twocolumn[1][]{#1}%
\maketitle
\begin{center}
    \centering
    \includegraphics[width=1.0\textwidth,height=5.2cm]{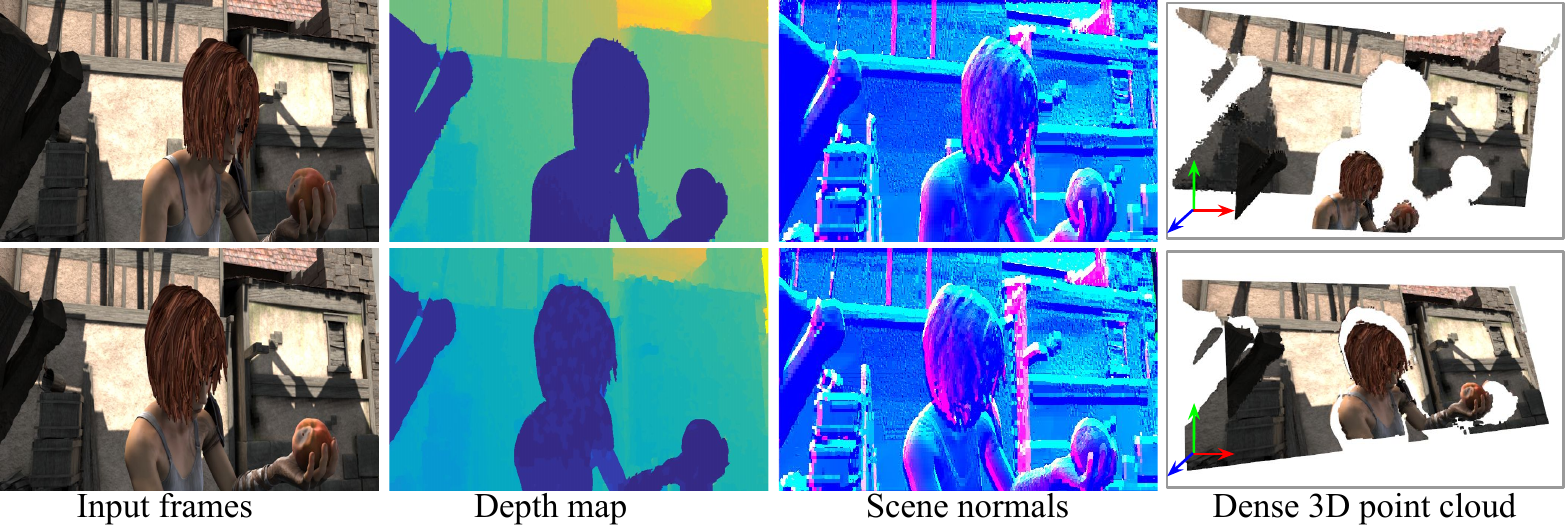}
    \captionof{figure}{Dense 3D reconstruction of a complex dynamic scene from two perspective frames using our method. Here, both the subject and the camera are moving with respect to each other. (MPI Sintel \cite{butler2012naturalistic}  alley\_1 frame 1 and 10).}
    \label{fig:introFig}
\end{center}%
}]

\begin{abstract}
This paper proposes a new approach for monocular dense 3D reconstruction of a complex dynamic scene from two perspective frames. By applying superpixel over-segmentation to the image, we model a generically dynamic (hence non-rigid) scene with a piecewise planar and rigid approximation.  In this way, we reduce the dynamic reconstruction problem to a ``3D jigsaw puzzle'' problem which takes pieces from an unorganized ``soup of superpixels''.  We show that our method provides an effective solution to the inherent \emph{relative scale ambiguity} in structure-from-motion. Since our method does not assume a template prior, or per-object segmentation, or knowledge about the rigidity of the dynamic scene, it is applicable to a wide range of scenarios. Extensive experiments on both synthetic and real monocular sequences demonstrate the superiority of our method compared with the state-of-the-art methods. 
\end{abstract}
%%%%%%%%% BODY TEXT
\section{Introduction}
Accurate recovery of dense 3D structure of dynamic scenes from images has many applications in motion capture \cite{loper2014mosh}, robot navigation\cite{geiger2013vision}, scene understanding \cite{Pollefeys-Reconstruction-Segmentation:CVPR-2013}, computer animation \cite{butler2012naturalistic} \etc. In particular, the proliferation of monocular camera in almost all modern mobile devices has elevated the demand for sophisticated dense reconstruction algorithm. When a scene is rigid, its 3D reconstruction can be estimated using conventional rigid-SfM (structure-from-motion) techniques \cite{hartley2003multiple}. However, real-world scenes are more complex containing not only rigid motions but also non-rigid deformations, as well as their combination. For example, a typical outdoor traffic scene consists of both multiple rigid motions of vehicles, and non-rigid motions of pedestrians \etc. Therefore, it is highly desirable to develop a {\em unified} monocular 3D reconstruction framework that can handle generic (complex and dynamic) scenes.

To tackle the problem of monocular 3D reconstruction for dynamic scenes, a straightforward idea is to first pre-segment the scene into different regions, each corresponding to a single rigidly moving object or a rigid part of an object, then apply rigid-SfM technique to each of the regions. This idea of object-level motion segmentation has been used in previous work for non-rigid reconstruction \cite{ranftl2016dense}\cite{Video-Popup:ECCV-2014}, and for scene-flow estimation \cite{menze2015object}. Russel \etal \cite{russell2014video} proposed to simultaneously segment a dynamic scene into its constituent objects and reconstruct a 3D model of the scene. Ranftl \etal \cite{ranftl2016dense} developed a two-stage pipeline (segmentation and then reconstruction) for monocular dynamic reconstruction. However, in a general dynamic setting, the task of densely segmenting rigidly moving objects or parts is not trivial. Consequently, inferring motion models for deforming shapes becomes very challenging. Furthermore, the success of object-level segmentation builds upon the assumption of multiple rigid motions, which fails to handle more general scenarios such as \eg when the objects themselves are nonrigid or deformable.

This motivates us to ask a natural question: ``{\it{Is object-level motion segmentation essential for the dense 3D reconstruction of a complex dynamic scene?}}''. In this paper, we will justify our stance by proposing an approach that is free from object-level motion segmentation. We develop a unified method that is able to recover a dense and detailed 3D model of a complex dynamic scene, from its two perspective images, without assuming motion types or segmentation. Our method is built upon two basic assumptions about the scene, which are: 1) the deformation of the scene between two frames is {\em locally-rigid,} but {\em globally as-rigid-as-possible}, 2) the structure of the scene in each frame can be approximated by a {\em piecewise planar}.  We call our new algorithm the {\em SuperPixelSoup} algorithm, for reasons that will be made clear in Section \ref{ss:algoOverview}. Fig-\ref{fig:introFig} shows some sample 3D reconstruction by our proposed method.

The main contributions of this work are:
\begin{enumerate}
\item We present a unified framework for dense two-frame 3D reconstruction of a complex dynamic scene, which achieves state-of-the-art performance.
\item We propose a new idea to resolve the inherent {\em relative scale ambiguity} for monocular 3D reconstruction by exploiting the as-rigid-as-possible ({\small ARAP}) constraint.
\end{enumerate}

\subsection{Related work}  
For brevity, we give a brief review only to previous works for monocular dynamic reconstruction that are mostly related to our work. The linear low-rank model has been used for dense nonrigid reconstruction. Garg \etal \cite{garg2013dense} solved the task with an orthographic camera model assuming feature matches across multiple frames. Fayad \etal \cite{piecewise-quadratic:ECCV-2010} recovered deformable surfaces with a quadratic approximation, again from multiple frames.  Taylor \etal \cite{taylor2010non} proposed a piecewise rigid solution using locally-rigid SfM to reconstruct a soup of rigid triangles. While their method is conceptually similar to ours, there are major differences: 1) We achieve \textit{two-view} dense reconstruction while they need multiple views(N $\geq$ 4); 2) We use the \textit{perspective camera model} while they rely on an orthographic camera model. Many real-world images such as a typical driving scene (\eg, KITTI) cannot be well explained  by orthographic projection; 3) We solve the relative scale indeterminacy issue, which is an inherent ambiguity for 3D reconstruction under perspective projection, while Taylor \etal's method does not suffer from this, at the cost of being restricted to the orthographic camera model.  Russel \etal \cite{russell2014video} and Ranftl \etal \cite{ranftl2016dense} used object-level segmentation for dense dynamic reconstruction. In contrast, our method is free from object segmentation, hence circumvents the difficulty associated with motion segmentation in a dynamic setting. The template-based approach is yet another method for deformable surface reconstruction. Yu \etal \cite{yu2015direct} proposed a direct approach to capturing dense, detailed 3D geometry of generic, complex non-rigid meshes using a single RGB camera. While it works for generic surfaces, the need of a template prevents its wider application to more general scenes.  Wang \cite{Template-free-3D-surface:ECCV-2016} introduced a template-free approach to reconstruct a poorly-textured, deformable surface. However, its success is restricted to a single deforming surface rather than the entire dynamic scene. Varol \etal \cite{varol2009template} reconstructed deformable surfaces based on a piecewise reconstruction, by assuming overlapping pieces.    

\begin{figure}[t] 
\begin{center}
\includegraphics[width=1.0\linewidth]{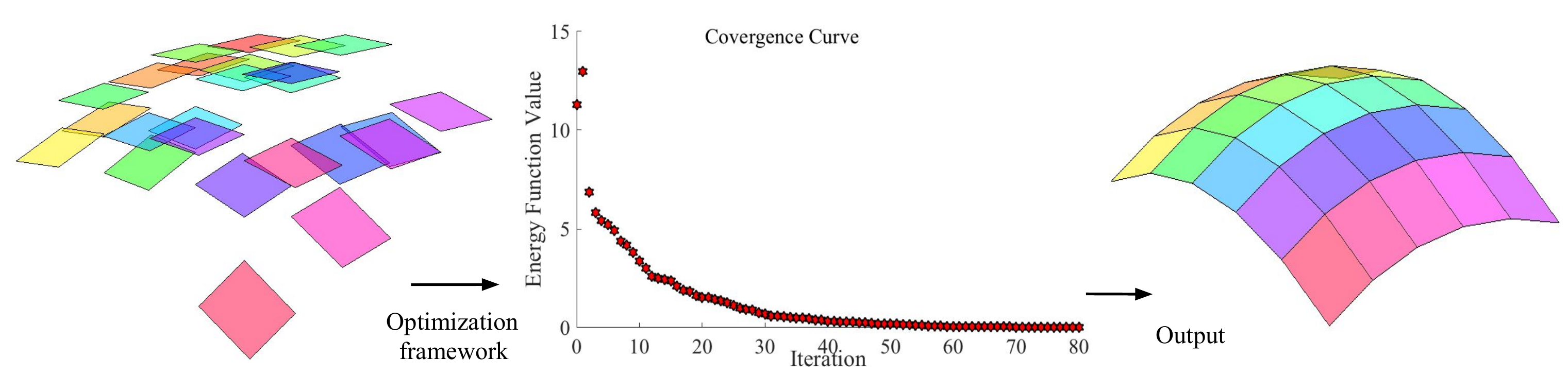}
\caption{Reconstructing a 3D surface from a soup of un-scaled superpixels via solving a 3D Superpixel Jigsaw puzzle problem.\label{fig:concept}}
\end{center}
\end{figure}
%==================================================================
\section{Overview of the proposed method} \label{ss:algoOverview}
% The value for each vertex and its edges are provided by variable initialization  \S \ref{ss:initialization} and energy function \S \ref{ss:EneryFunction}.
\begin{figure*}
\centering
\includegraphics[width=1.0\textwidth] {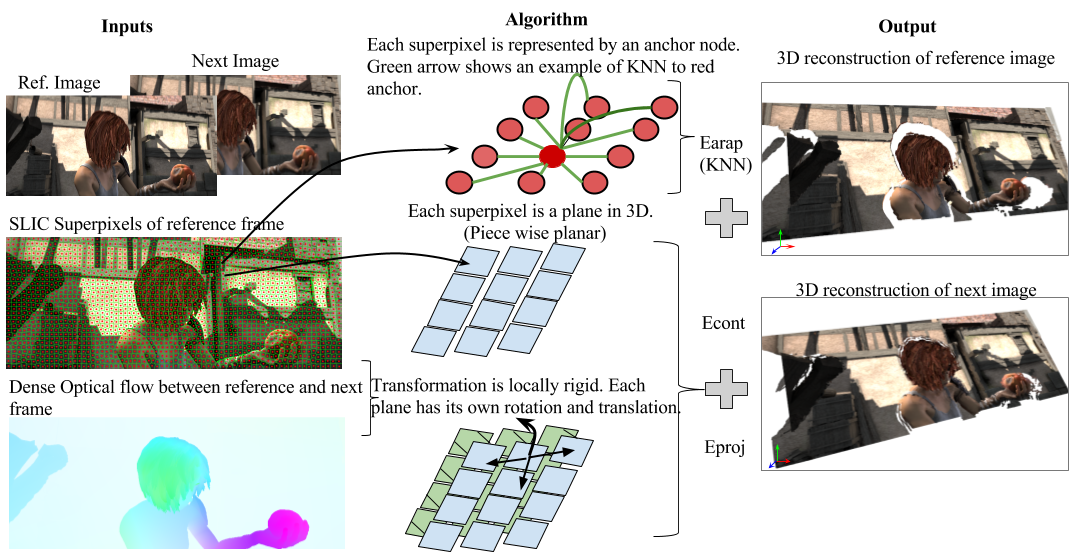}~~~
\caption{Flow diagram of the proposed approach. {\bf{Left column:}} The inputs for our algorithm a) Two input frames b) SLIC superpixels \cite{achanta2012slic} of the reference frame c) Dense optical flow between two frames. {\bf{Middle column:}} Each individual superpixel is represented by an anchor node (in dark red). Every anchor node constrains the motion of $K$ other anchor node ($E_{\mathrm{arap} }$) in both frames. The depth continuity term $(E_{\mathbf{cont} })$ is defined only for neighboring superpixels that shares the common boundary. {\bf{Right column:}} The dense 3D point clouds of the reference frame and the next frame, where each individual plane in the next frame is related to the reference frame via a rigid motion.}
\label{fig:FlowDiagram}
\end{figure*}

In this section, we present a high-level overview of our ``SuperPixel Soup'' algorithm for dense 3D scene reconstruction of a complex dynamic scene from two frames.

Given two perspective images (denoted as the reference image $\mathbf{I}$ and the next image $\mathbf{I}'$) of a generally dynamic scene, our goal is to recover the dense 3D structure of the scene.  We first pre-segment the image into superpixels, then model the deformation of the scene by the union of piece-wise rigid motions of its superpixels. Specifically, we divide the overall non-rigid reconstruction into small rigid reconstruction for each individual superpixel, followed by an assembly process which glues all these local individual reconstructions in a globally coherent manner.  While the concept of the above divide-and-conquer procedure looks simple, there is, however, a fundamental difficulty (of {\em relative scale indeterminacy}) in its implementation. {Relative scale indeterminacy} refers to the well-known fact that using a moving camera one can only recover the 3D structure up to an unknown scale.  In our method, the individual rigid reconstruction of each superpixel can only be determined up to an unknown scale, the assembly of the entire non-rigid scene is only possible if and only if these relative scales among the superpixels are solved --which is, however, a challenging open task itself. 
%{\em Note:} whenever we say ``scale-indeterminacy'' we actually mean the ambiguity in ``relative scales".

In this paper, we show how this can be done, under two very mild assumptions (about the dynamic scene and about the deformation).  Specifically, these assumptions are:
\begin{itemize}[noitemsep]
\item {Basic Assumption-1: The transformation (\ie deformation) between the two frames are locally {\em piecewise-rigid}, and globally {\bf as rigid as possible}. In other words, the deformation is not arbitrary but rather regular in terms of rigidity.}
\item {Basic Assumption-2: The 3D scene surface to be reconstructed is {\bf piecewise-smooth} (or moreover, {\bf piecewise-planar}) in both frames.}  
\end{itemize} 
Under these assumptions, our method solves the unknown relative scales and obtains a globally-coherent dense 3D reconstruction of a complex dynamic (hence generally nonrigid) scene from its two perspective views. 

Intuitively, our new method can be understood as the following process: Suppose every individual superpixel corresponds to a small planar patch moving rigidly in 3D space. Since the correct scales for these patches are not determined, they are floating in 3D space as a set of unorganized superpixel soup.  Our method then starts from finding for each superpixel an appropriate scale, under which the entire set of superpixels can be assembled (glued) together coherently, forming a piecewise smooth surfaces, {\em as if} playing the game of ``3D jig-saw puzzle''.  Hence, we call our method the ``SuperPixel Soup'' algorithm (see Figure \ref{fig:concept} for a conceptual visualization).

The overall procedure of our method is presented in Algorithm-1.
\begin{algorithm}[h!]
\label{Algorithm}
\caption{:~~SuperPixel Soup}
\begin{algorithmic}
\STATE {\bf{Input:}} Two monocular image frames and dense optical flow correspondences between them.
\STATE {\bf{Output:}} 3D reconstruction of both image. 
 \STATE 1. Divide the image into $N$ superpixel and construct a K-NN graph to represent the entire scene as a graph $G(V, E)$ defined over superpixels \S \ref{ss:PF}.
 \STATE 2. Employ the two-view epipolar geometry to recover the rigid motion and 3D geometry for each 3D superpixel.
\STATE 3. Optimize the proposed energy function to assemble (or glue) and align all the reconstructed superpixels (``3D Superpixel Jigsaw Puzzle'').

\end{algorithmic}
\end{algorithm}

%=============================================================
%\section{Approach} \label{ss:TP} 
\section{Problem Statement}
To implement the above idea of piecewise rigid reconstruction, we first partition the reference image $\mathbf{I}$ into superpixels $\{\mathbf{s}_1, \mathbf{s}_2,.., \mathbf{s}_{i},.., \mathbf{s}_{N}\}$, where each superpixel $\mathbf{s}_{i}$ is parametrized by its boundary pixels $\{\mathbf{x}_{bi}=[u_{bi}, v_{bi},1]^{T} ~| b = 1,..., B_i\}$ in the image plane. We further define an {\em anchor point} $\mathbf{x}_{ai}$ for each superpixel, as the centroid point of the superpixel. Such a superpixel partition of the image plane naturally induces a piecewise planar segmentation of the corresponding 3D scene surface.  We call each of the 3D segments as a 3D superpixel, and denote its boundary coordinates (in 3D space) as $\{\mathbf{S}_{i}\}$ in capital $\mathbf{S}$. Although {\it{surfel}} is perhaps a better term, we nevertheless call it ``3D superpixel'' for the sake of easy exposition. We further assume each 3D superpixel is a small 3D {\em planar patch}, parameterized by surface normal ${\mathbf{n}_{i}}\in\mathbb{R}^3$, 3D anchor-point ${\mathbf{X}_{ai}}$, and 3D  boundary-points $\{ {\mathbf{X}_{bi}\}}$  (\ie these are the pre-images of $\mathbf{x}_{ai}$ and $\{\mathbf{x}_{b i}\}$). We assume every 3D superpixel $\mathbf{s}_i$ moves rigidly according $\mathbf{M}_{i}=
\left(
 \begin{smallmatrix} 
   \mathbf{R}_{i} & \lambda_{i} \hat{\mathbf{t}}_{i}\\
   \mathbf{0} & 1
 \end {smallmatrix} 
\right) \in\mathrm{SE}(3),$
where $\mathbf{R}_i$ represents rotation, $\hat{\mathbf{t}}_{i}$ is the translational direction, and $\lambda_{i}$ is the unknown scale.

Now we are in a position to state the problem in a more precise way:  Given two intrinsically calibrated perspective images $\mathbf{I}$ and $\mathbf{I}'$ of a generally dynamic scene and the corresponding dense correspondences, \ie, optical flow field, our task is to reconstruct a piecewise planar approximation of the dynamic scene surface. We need a {\em dense} flow field, but do not require it to be perfect because it is only used to initialize our algorithm, and as the algorithm runs, the final flow field will be refined. The deformable scene surface in the reference frame (\ie, $\mathbf{S}$) and the one in the second frame (\ie, $\mathbf{S}'$) are parametrized by their respective 3D superpixels $\{\mathbf{S}_{i}\}$ and $\{\mathbf{S}'_{i}\}$, where each ${\mathbf{S}}_{i}$ is described by its surface normal $\mathbf{n}_{i}$ and an anchor point $\mathbf{X}_{ai}$. Any 3D plane can be determined by an anchor point $\mathbf{X}_{ai}$ and a surface normal $\mathbf{n}_{i}$. If one is able to estimate all the 3D anchor points and all the surface normals, the problem is solved.
%==========================================================

%========================================================
\section{Solution}\label{ss:PF}

\paragraph{Build a K-NN graph.} We identify a 3D superpixel by its anchor point. The distance between two 3D superpixels is defined as the Euclidean distance between their anchor points in 3D space.  

\begin{figure*}
\centering
\includegraphics[width=1.0\textwidth] {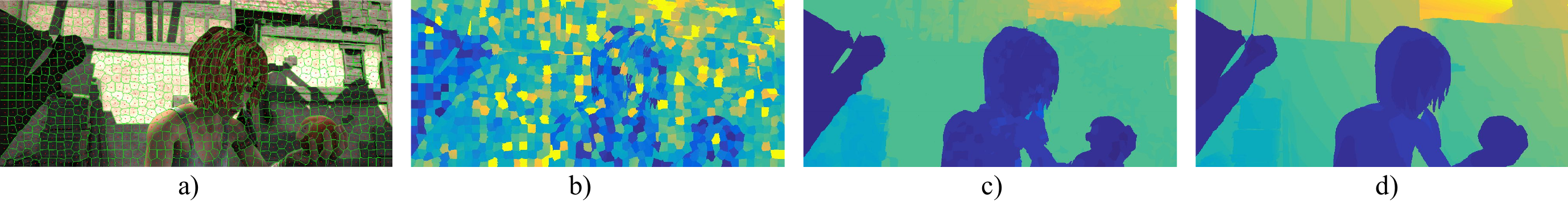}~~~
\caption{ a) Superpixelled reference image b) Individual superpixel depth with arbitrary scale ({\it{unorganised superpixel soup}}) c) recovered depth map using our approach ({\it{organised superpixel soup}}) d) ground-truth depth map.}
\label{fig:solutionExample}
\end{figure*}

By connecting K nearest neighbors, we build a K-NN graph G(V,E) (\eg as illustrated in Fig.~\ref{fig:FlowDiagram} and Fig.~\ref{fig:solutionExample}).  The graph vertices are anchor points, connecting with each other via graph edges. Overloading notation, we let $\mathbf{X}_{ai}=[{X}_{ai}, Y_{ai}, Z_{ai}]^{T}$ represent 3D world coordinates of the $i$-th superpixel. Suppose that we know the perfect $\mathbf{M}_{i}$, $\mathbf{n}_i$ for each individual $\mathbf{S}_{i}$, then $\mathbf{S}$ can be mapped to $\mathbf{S}'$ by moving each individual superpixel based on its corresponding locally rigid motion.  The world and the image coordinates in the subsequent frames can be inferred by $\mathbf{X}_{ai}'= \mathbf{M}_i\mathbf{X}_{a i}$ and $\mathbf{s}_i'=\mathbf{K}\left(\mathbf{R}_i-\frac{\mathbf{t}_i\mathbf{n}_i^T}{d_i}\right)\mathbf{K}^{-1}\mathbf{s}_i$, where the latter represents a plane-induced homography \cite{hartley2003multiple}, with ${d_i}$ as the depth of the plane.

%===============================================
\paragraph{As-Rigid-As-Possible (ARAP) Energy Term.} Our new method is built upon the idea that the correct scales of 3D superpixels can be estimated by enforcing prior assumptions that govern the deformation of the dynamic surface.  Specifically, we require that, locally, the motion that each 3D-superpixel undergoes is rigid, and globally the entire dynamic scene surface must move as-rigid-as-possible (ARAP). In other words, while the dynamic scene is globally non-rigid, its deformation must be {\em regular} in the sense that it deforms as rigidly as possible.  To implement this idea, we define an ARAP-energy term as:
\begin{equation}\label{eq:EARAP}
{\begin{aligned}
& \displaystyle E_{\textrm{arap}} = \sum_{i=1}^{N} \sum_{k \in \mathcal{N}_i} w_{1}(\mathbf{s}_{ai}, \mathbf{s}_{a k}) \|\mathbf{M}_{i}- \mathbf{M}_{k}\|_F + \\
& \displaystyle w_{2}(\mathbf{s}_{ai}, \mathbf{s}_{a k}).\Big| \|{\mathbf{X}}_{ai}- {\mathbf{X}}_{ak}\|_2 - \|{\mathbf{X}}'_{ai}-{\mathbf{X}}'_{ak}\|_2\Big|_1.
\end{aligned}}
\end{equation}
%w_{0} \|\mathbf{M}_{i}- \mathbf{M}_{k}\|_2 + \\
Here, the first term favors smooth motion between local neighbors, while the second term encourages inter-node distances between the anchor node and its K nearest neighbor nodes (denoted as $k \in$ $\mathcal{N}_i$) to be preserved before and after motion (hence as-rigid-as-possible). We define the weighting parameters as:
\begin{equation}\label{eq:wc} w_{1}(\mathbf{s}_{ai}, \mathbf{s}_{a k}) = w_{2}(\mathbf{s}_{ai}, \mathbf{s}_{a k}) = \exp(- \beta \|\m s_{a i} - \m s_{a k} \|).
\end{equation}
These weights are set to be inversely proportional to the distance between two superpixels. This is to reflect our intuition that, the further apart two superpixels are, the weaker the $E_{\textrm{arap}}$ energy is. Although there may be redundant information in these two terms, we keep both nonetheless for the sake of flexibility in algorithm design. Note that, this term is only defined over anchor points, hence it enforces no depth smoothness along boundaries. The weighting term in $E_{\textrm{arap}}$ advocates the local rigidity by penalizing over the distance between anchor points. This allows immediate neighbors to have smooth deformation over time. Also, note that $E_{\textrm{arap} }$ is generally {\em non-convex}. 
\\
\\

{\bf{Planar Re-projection Energy Term.}} With the assumption that each superpixel represents a plane in 3D, it must satisfy corresponding planar reprojection error in 2D image space. This reprojection cost reflects the average dissimilarity in the optical flow correspondences across the entire superpixel due to its motion. Therefore, it helps us to constrain the surface normals, rotation and translation direction such that they obey the observed planar homography in the image space.
\begin{equation}\label{eq:planarerror}
\begin{aligned}
& \displaystyle E_{\mathrm{proj}} = \sum_{i=1}^{N} \frac{w_{3}}{|\mathbf{s}_i|}\sum_{{j}=1}^{|\mathbf{s}_i|}\|(\mathbf{s}_{i}^{j})'- \mathbf{K}\left(\mathbf{R}_{i}-\frac{\mathbf{t}_i \mathbf{n}_{i}^{T}}{ d_{ i}}\right)\mathbf{K}^{-1} (\mathbf{s}_{i}^j) \|_F.
\end{aligned}
\end{equation}
where $|\mathbf{s}_i|$ represents the total number of pixel inside the $i^{th}$ superpixel \footnote{For brevity, we slightly abuse notation; both terms in Eq:-\ref{eq:planarerror} represent inhomogeneous image coordinate.}.
\\
\\
{\bf{3D Continuity Energy Term.}} To favor a continuous/smooth surface reconstruction, we require two neighboring superpixels to have a smooth transition at their boundaries. We define a 3D continuity energy term as:
\begin{equation}\label{eq:Econt}
\begin{aligned}
& \displaystyle E_{\mathrm{cont}} = 
\sum_{i=1}^{N} \sum_{k \in \mathcal{N}_i}  w_4(\mathbf{s}_{bi}, \mathbf{s}_{bk}) \Big(\|\mathbf{X}_{bi}- \mathbf{X}_{bk}\|_F  + \\
& \displaystyle \rho(\|\mathbf{X}'_{bi}-\mathbf{X}'_{bk}\|_F)\Big). 
\end{aligned}
\end{equation}
This term ensures the 3D coordinates across superpixel boundaries to be continuous in both frames. The neighboring relationship in $E_{\mathrm{cont}}$ is different from $E_{\mathrm{arap}}$ term. Here, the neighbors share common boundaries with each other. For each boundary pixel of a given superpixel, we consider its 4-connected neighboring pixels. $w_4$ is a trade-off scalar, which is defined as:
\begin{equation}
w_{4}(\mathbf{s}_{bi}, \mathbf{s}_{bk}) = \exp(-\beta \|\m I(\mathbf{s}_{bi}) - \m I(\mathbf{s}_{bk}) \|_F), 
\end{equation}\ie weighting the inter-plane transition by the color difference. Here, subscript '$bi$' and '$bk$' indicate that the involved pixels shares the common boundary ('$b$') between $i^{th}$ and $k^{th}$ superpixel in the image space. $\rho$ is a truncation function defined as $\rho=\min(., \sigma)$ to allow piecewise discontinuities. Here, $\beta$ is a trade-off constant chosen empirically.

\paragraph{Combined Energy Function.} Recall that our goal is to estimate piecewise rigid motion ($\mathbf{R}_i, \mathbf{t}_i$), depth $d_i$, surface normal $\mathbf{n}_i$ and scale $\lambda_i$ for each planar superpixel in 3D, given initialization. The key is to estimate the unknown relative scale $\lambda_i$.  We solve this by minimizing the following energy function $E = E_{\mathrm{arap}} + \alpha_1 E_{\mathrm{proj}} + \alpha_2 E_{\mathrm{cont} }$, namely,
\begin{equation}
\begin{aligned}
& \displaystyle \underset{{\lambda_i}, {\mathbf{n}_i}, {d_i}, \mathbf{R}_i, \mathbf{t}_i} \min  \hspace{.1cm} E = E_{\mathrm{arap}} + \alpha_1 E_{\mathrm{proj}} + \alpha_2 E_{\mathrm{cont}},  \\
& \displaystyle \text{{s. t.}} \displaystyle{\sum_{i=1..N}\lambda_{i}= 1, \lambda_{i}>0}.
\end{aligned}\label{eq:CostFunction}
\end{equation}  The last equality constraint fixes the unknown freedom of a global scale. $\lambda_i > 0$ enforces the cheriality constraint \cite{hartley2003multiple}. 

\paragraph{Optimization.}  
The above energy function (Eq.- \ref{eq:CostFunction}) is non-convex. We first solve the relative scales $\lambda_i$ efficiently by minimizing the ARAP term in Eq.-\eqref{eq:EARAP} using interior-point methods \cite{benson2002interior}.  Although the solutions found by the interior point method are at best local minimizers, empirically they appear to give good 3D  reconstructions.  In our experiments, we initialized all $\lambda_i$ with an initial value of $\frac{1}{N}$. 

Assigning superpixels to a set of planes can lead to non-smooth blocky effect at their boundaries. To smooth these blocky effects, we employ a refinement step to optimize over the surface normals, rotations, translations, and depths for all 3D superpixels using  Eq.- \ref{eq:planarerror} and Eq.-\ref{eq:Econt}. We solved the resultant discrete-continuous optimization with  the Max-Product Particle Belief propagation (MP-PBP) procedure by using the TRW-S algorithm \cite{kolmogorov2006convergent}. In our implementation, we generated 50 particles as proposals for the unknown parameters. Repeating the above strategy for 5-10 iterations, we obtained a smooth and refined 3D structure of the dynamic scene.

\paragraph{Implementation details.} We partitioned a reference image into about 1,000-2,000 superpixels \cite{achanta2012slic}.  We used a state-of-the optical flow algorithm \cite{bailer2015flow} to compute dense correspondences across two frames. Parameters like $\alpha_1$, $\alpha_2$, $\beta$, $\sigma$ were tuned differently for different datasets. However, $\beta = 3$ and $\sigma = 15$ are fixed for all our tests on MPI Sintel and on VKITTI.  To initialize the iteration, local rigid motion is estimated using traditional SfM pipeline \cite{hartley2003multiple}. Our current implementation in C++/MATLAB takes around 10-12 minutes to converge for images of size $1024\times 436$ on a regular desktop with Intel core i7 processor.

%================================================================================
\section{Experiments}
\begin{figure*}[h!]
\centering
\includegraphics[width=1.0\textwidth] {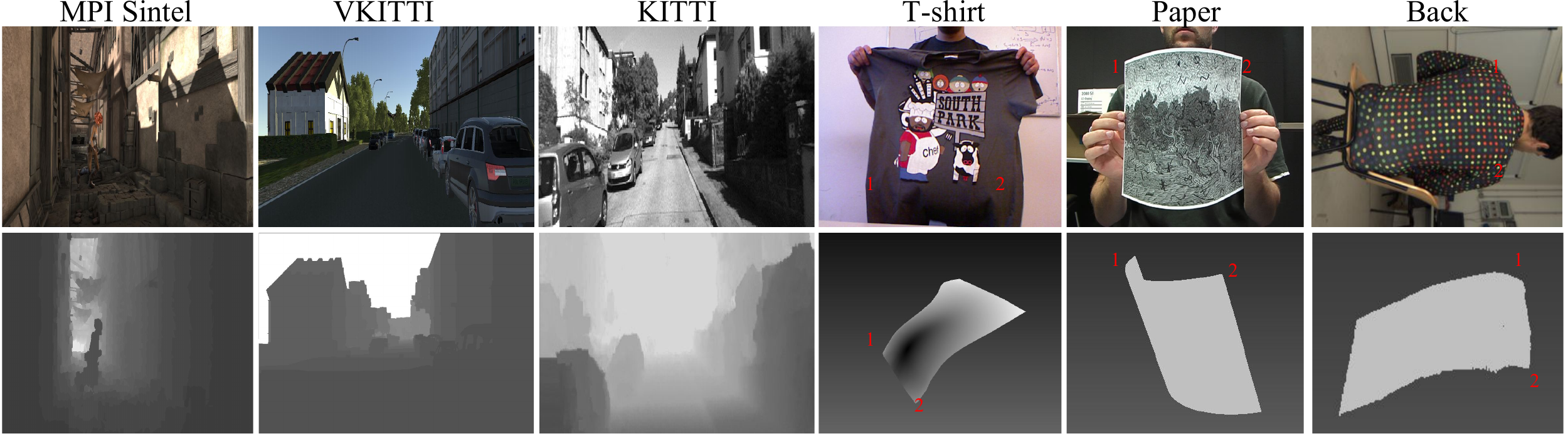}~~~
\caption{\small 3D reconstruction and depth map obtained using our algorithm on different benchmarking datasets. The first three columns demonstrate the reconstruction of the entire scene that is composed of rigid and complex motion. The last three columns show the accurate reconstruction of deformable objects on real non-rigid benchmark datasets.}
\label{fig:resulteachCategory}
\end{figure*}

We evaluated the performance of our method both qualitatively and quantitatively on various bedatasets that contain dynamic objects: the KITTI dataset \cite{geiger2013vision}, the virtual KITTI \cite{gaidon2016virtual}, the MPI Sintel \cite{butler2012naturalistic} and the YouTube-Objects \cite{prest2012learning}. We also tested our method on some commonly used non-rigid deformation data: {\it{Paper}}, {\it{T-shirts}} and {\it{Back}} sequence \cite{varol2009template} \cite{varol2012constrained}\cite{garg2013dense}. Example images and our reconstruction results are illustrated in Fig.~\ref{fig:resulteachCategory}. 

%Change the order of the figures to consist with the text... 

{\textbf{Evaluation Metrics:}}
For quantitative evaluation, the errors are reported in \ie mean relative error (MRE), defined as $\frac{1}{P}\sum_{i=1}^{P}|z_{gt}^{i} - z_{est}^{i}|/ z_{gt}^{i}$. Here, $z_{est }^{i}$,  $z_{gt}^{i}$ denotes the estimated, and ground-truth, depth respectively with $P$ as the total number of 3D points.  The error is computed after re-scaling the recovered shape properly, as the reconstruction is only made up to an unknown global scale.  We used MRE for the sake of consistency with previous work \cite{ranftl2016dense}.  Quantitative evaluations for the YouTube-Objects dataset and the Back dataset are missing because for them no ground-truth results are provided.

{\textbf{Baseline Methods:}} The performance of our presented method is compared to several monocular dynamic reconstruction methods, which include the Block Matrix Method (BMM) \cite{dai2014simple}, Point Trajectory Approach (PTA) \cite{akhter2011trajectory}, and Low-rank Reconstruction (GBLR) \cite{fragkiadaki2014grouping}), Depth Transfer (DT) \cite{karsch2014depth}, and (DMDE) \cite{ranftl2016dense}. \footnote{We did not compare our method with \cite{russell2014video} due to the code provided by the authors of \cite{russell2014video} crashed unexpectedly on several of the test sequences.} 

\begin{figure*}[h!]
\centering
\includegraphics[width=1.0\textwidth] {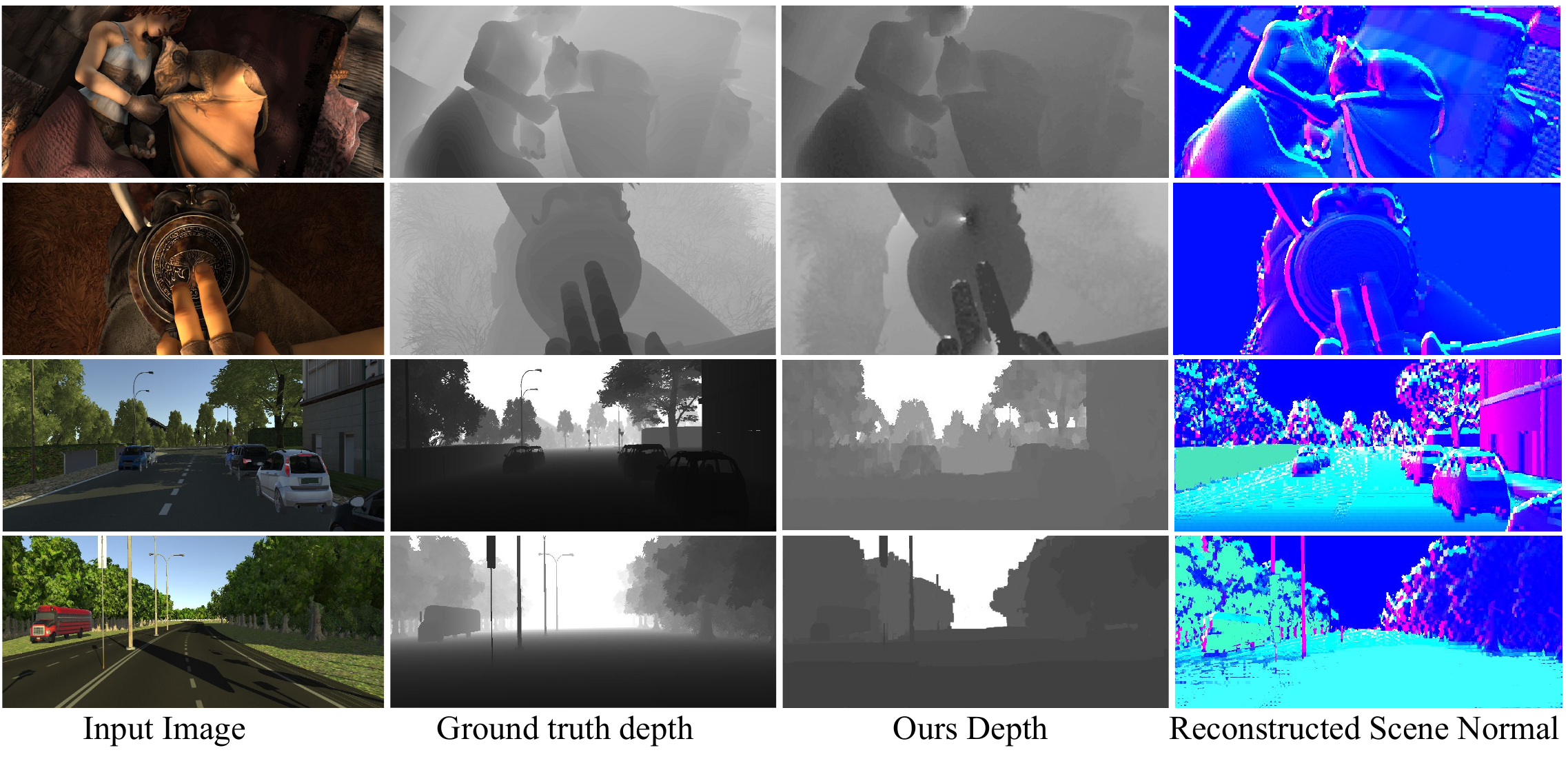}~~~
\caption{\small Depth map and scene normals on MPI and VKITTI dataset. }
\label{fig:resulteachCategory2}
\end{figure*}

In Fig-(\ref{fig:resulteachCategory2}) we show the recovered depth map along with scene surface normals. These results highlight the effectiveness of our method in handling diverse scenarios. 
\begin{figure*}[h!]
\centering
\includegraphics[width=1.0\textwidth] {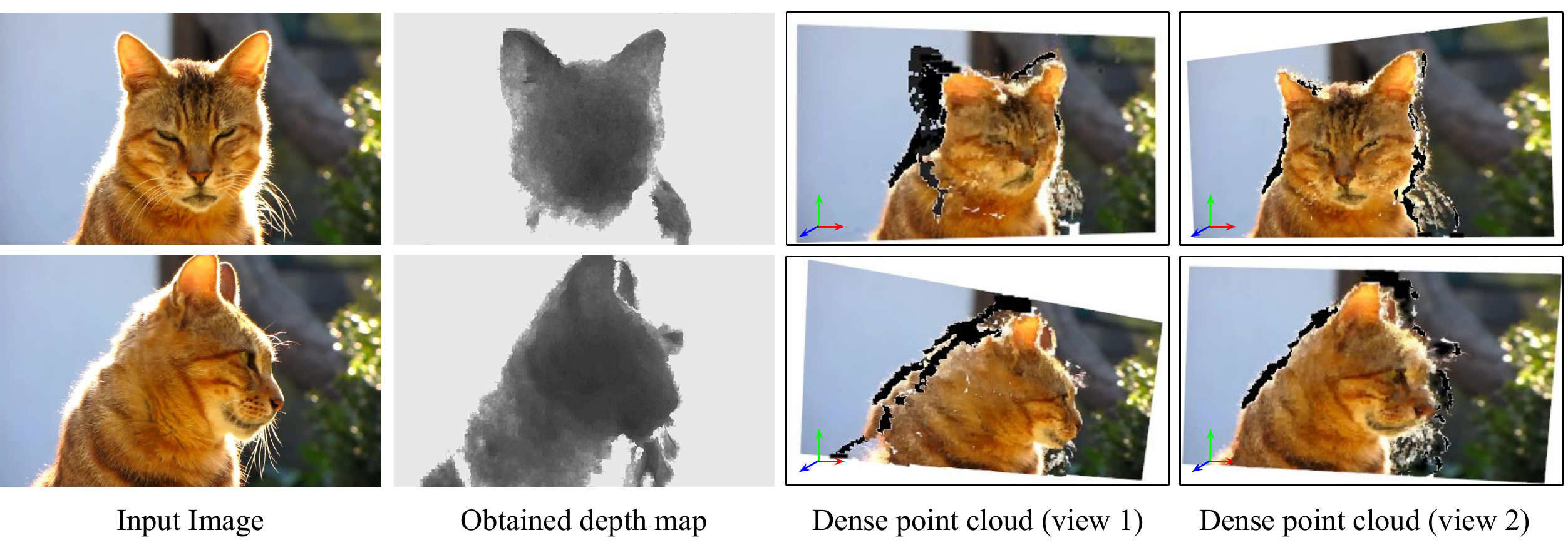}~~~
%\caption{Depth and 3D reconstruction results for a cat sequence taken from Youtube-Objects Dataset\cite{prest2012learning}.}
\caption[The LOF caption]{\small{Depth and 3D reconstruction results for the cat sequence taken from YouTube-Objects Dataset\cite{prest2012learning}\footnotemark}. For this experiment, we used 10,000 superpixels.}
\label{fig:resulteachCategory3}
\end{figure*}

\begin{figure*}[!htp]
  \begin{center}
  \subfigure[\label{fig:MPIVKITTIStats}]{\includegraphics[width=0.49\linewidth ]{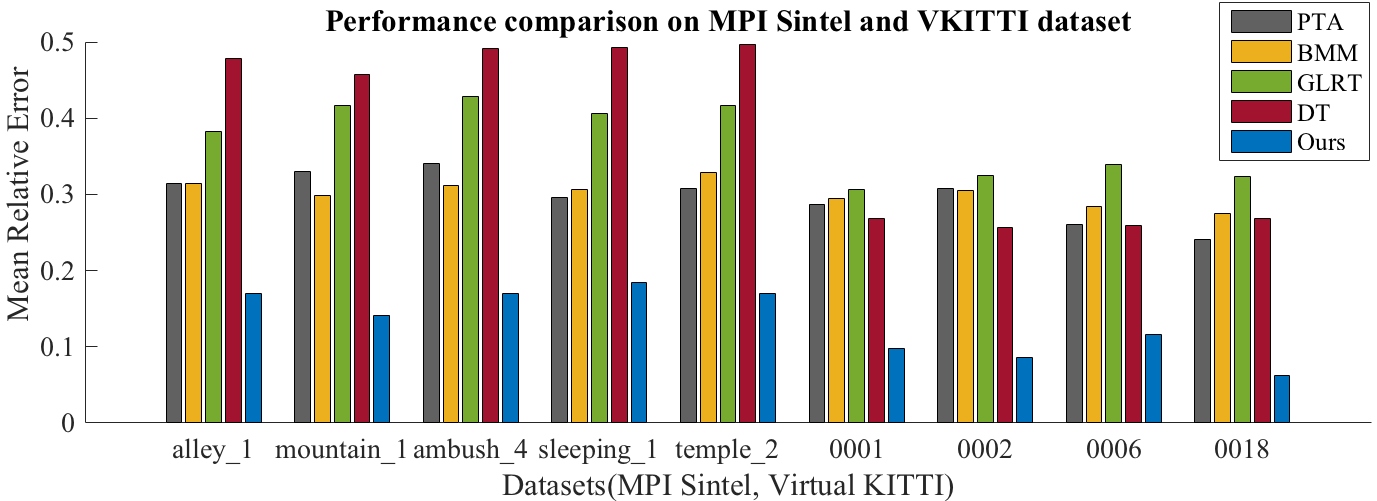}}
  \subfigure[\label{fig:KITTIStats}]{\includegraphics[width=0.49\linewidth, height=0.18\linewidth]{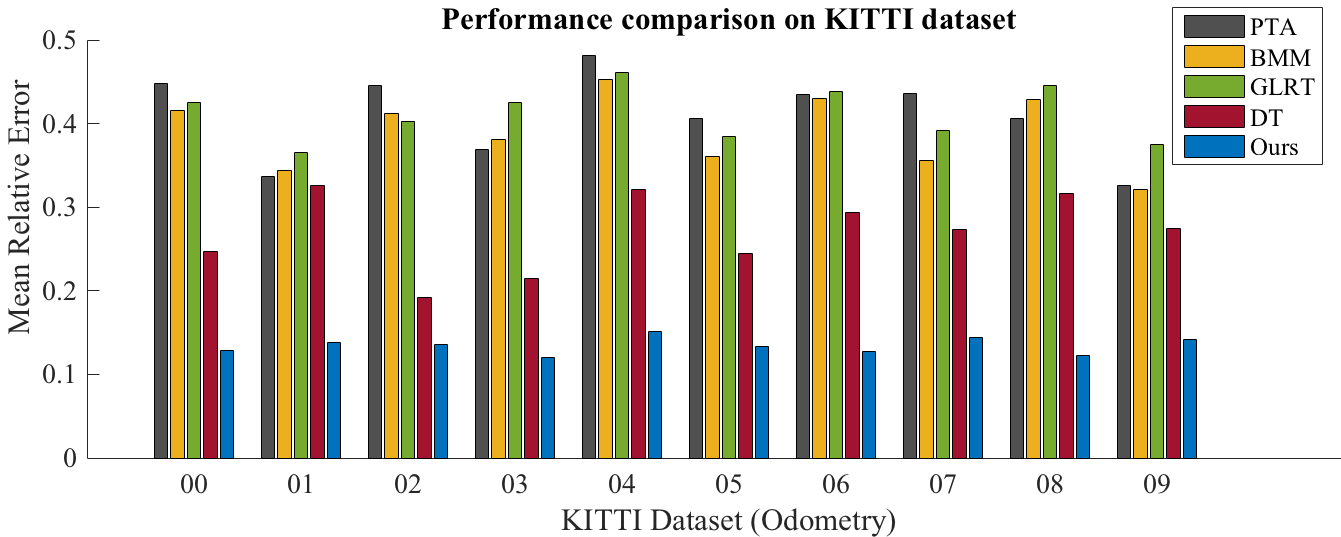}}
\end{center}
  \caption{\small Quantitative evaluation on benchmark datasets. The depth error is calculated by adjusting the numerical scale of obtained depth map to ground-truth value, to account for global scale ambiguity. (a)-(b) comparison on MPI, Virtual KITTI and  KITTI dataset. PTA \cite{akhter2011trajectory}, BMM \cite{dai2014simple}, GLRT\cite{fragkiadaki2014grouping}, DT \cite{karsch2014depth}. These numerical values show the fidelity of reconstruction that can be retrieved on benchmark datasets using our formulation.}
  \label{fig:comparison_vkitti_kitti_mpi}
\end{figure*}

{\bf{MPI Sintel}:} This dataset is derived from an animation movie with complex dynamic scenes. It contains highly dynamic sequences with large motions and significant illumination changes.  It is a challenging dataset particularly for the piece-wise planar assumption due to the presence of many small and irregular shapes in the scene. 
We selected 120 pairs of images to test our method, which includes alley\_1,  ambush\_4, mountain\_1,  sleeping\_1 and temple\_2.  Fig-\ref{fig:MPIVKITTIStats} gives quantitative comparisons against several other competing methods. As observed in the figure, our method outperforms all the competing methods on all the testing sequences shown here.

{\bf{Virtual KITTI}:}
The Virtual KITTI dataset contains computer rendered photo-realistic outdoor driving scenes which resemble the KITTI dataset. The advantage of using this dataset is that it provides perfect ground-truths for many measurements.  Furthermore, it helps to simulate algorithm related to dense reconstruction with noise free and distortion-free images, facilitating quick experimentation. We selected 120 images from 0001\_morning, 0002\_morning, 0006\_morning and 0018\_morning. The results obtained are shown in Figure \ref{fig:MPIVKITTIStats}. Again, our method outperforms all the competing methods with a clear margin on all the test sequences.
\begin{table*}[h!]
\centering
\begin{tabular}{c|c|c|c|c|c|c}
\hline
\begin{tabular}[c]{@{}l@{}}~~~\textbf{Method $\rightarrow$} \\ (Method type)\end{tabular} & \begin{tabular}[c]{@{}l@{}}~~~DT \cite{karsch2014depth}\\ (Single frame)\end{tabular} & \begin{tabular}[c]{@{}l@{}}~GLRT \cite{fragkiadaki2014grouping}\\ (Multi-frame)\end{tabular} & \begin{tabular}[c]{@{}l@{}}~BMM \cite{dai2014simple}\\ (Multi-frame)\end{tabular} & \begin{tabular}[c]{@{}l@{}}~~PTA \cite{akhter2011trajectory}\\ (Multi-frame)\end{tabular} & \begin{tabular}[c]{@{}l@{}}DMDE \cite{ranftl2016dense}\\ (Two-frame)\end{tabular} & \begin{tabular}[c]{@{}l@{}}~~~~~Ours\\ (Two-frame)\end{tabular} \\ \hline
MPI Sintel & 0.4833 & 0.4101  & 0.3121 & 0.3177 & 0.297  & {\bf{0.1669}} \\ \hline
Virtual KITTI & 0.2630 & 0.3237 & 0.2894 & 0.2742 & - & {\bf{0.1045}} \\ \hline
KITTI & 0.2703 & 0.4112 & 0.3903 & 0.4090 & 0.148 & {\bf{0.1268}} \\ \hline
kinect\_paper & 0.2040 & 0.0920 & {\bf{0.0322}} & 0.0520 & - & 0.0476 \\ \hline
kinect\_tshirt & 0.2170 & 0.1030 & 0.0443 & {\bf{0.0420}} & - & 0.0480 \\ 
\hline
\end{tabular}
\caption{\small Performance Comparison: This table lists the MRE errors. For DMDE \cite{ranftl2016dense} we used its previously reported result as its implementation is not available publicly.}
\label{tab:comparisonOverall}
\end{table*}

{\bf{KITTI}:} We tested real KITTI to evaluate our method's performance for noisy real-world sequences.  We used the KITTI's sparse LiDAR points as the 3D ground-truth for evaluation. We also used other sequences for qualitative analysis (see Figure \ref{fig:resulteachCategory}). Figure \ref{fig:KITTIStats} demonstrates the obtained depth accuracy. Our method achieves the best performance for all the testing sequences.

\begin{figure*}[h!]
\centering
\includegraphics[width=1.0\textwidth, height=0.2\textwidth] {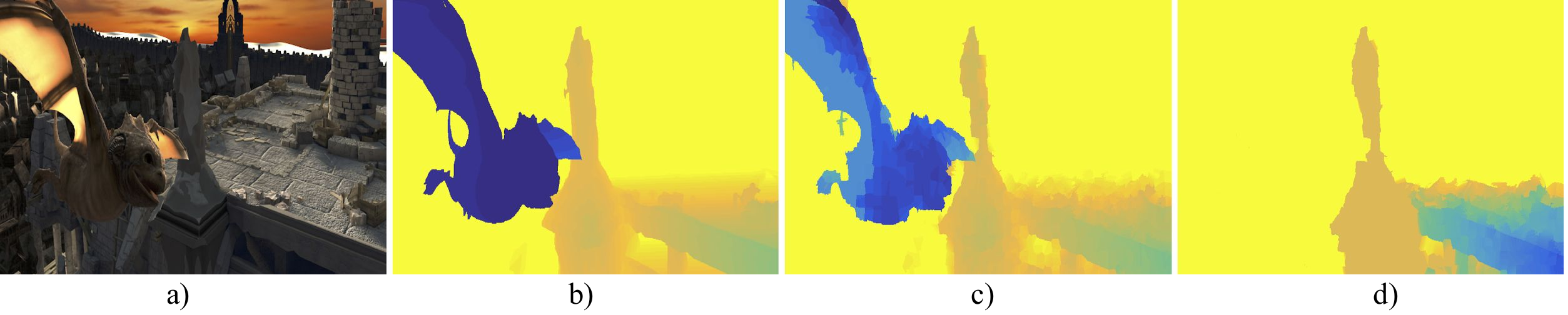}~~~
\caption{\small Effect of parameter K in building the K-NN graph.  Our algorithm results in good reconstruction if a suitable K is chosen, in accordance with levels of complexity of the dynamic scene. b) Ground-truth depth-map (scaled for illustration purpose). c) when K=4, a reasonable reconstruction is obtained. d) when K=20, regions tend to grow bigger. (Best viewed in color.)}
\label{fig:dragonfailure}
\end{figure*}
%========================================================
{\bf{YouTube-Objects:}} We tested our method on sequences from the Youtube-Objects Dataset \cite{prest2012learning}. These are community-contributed videos downloaded from the YouTube. Due to the lack of ground truth 3D reconstruction, we only show the results in Fig.~\ref{fig:resulteachCategory3} visually. 

{\bf{Non-rigid datasets (Paper, T-shirt, Back)}:}
%It is well known that recovering 3D deformation without physics model or template \cite{salzmann2008local} for non-rigid surfaces is a very challenging task. 
We benchmarked our method in commonly used deformable object sequences, namely, Kinect\_Paper and Kinect\_Tshirt \cite{varol2012constrained}.  Table-\ref{tab:comparisonOverall} presents the mean depth error obtained on these sequences. Note that all the benchmarking non-rigid structure-from-motion methods reported in Table-\ref{tab:comparisonOverall} (GLRT \cite{fragkiadaki2014grouping}, BMM \cite{dai2014simple}, and PTA \cite{akhter2011trajectory}) used multi-frame while our method only used two frames. Qualitative results are demonstrated in Fig.~\ref{fig:resulteachCategory}.
  
{\bf{Comparison:}}  Table \ref{tab:comparisonOverall} provides a statistical comparison between our method and other competing methods.  It shows that our method delivers consistently superior reconstruction accuracy on these benchmarking datasets, even better than those methods which use multiple image frames. 
\footnotetext{Intrinsic matrix was obtained through personal communication.}
\footnotetext{Intrinsic matrix for the Back sequence is not available with dataset. We made an approximate estimation.}
\textbf{Effect of K:} Under our method, the ARAP energy term is evaluated within K nearest neighbors, different K may have a different effect on the resultant 3D reconstruction. We conducted an experiment to analyze the effect of varying K on the MPI Sintel dataset and the results are illustrated in Fig.~\ref{fig:dragonfailure}. With the increase of K, the recovered scene becomes more rigid, as the neighborhood size increases. When k=20, the dragon region was absorbed into the sky region, which results in an incorrect reconstruction. In most of our experiments, we used a K in the range of $15-20$, which achieved satisfactory reconstructions.

Our approach may disappoint if the neighboring relations between superpixels do not hold in the successive frame due to the substantial motion. A couple of examples for such situations are discussed and shown in the supplementary material for better understanding. Furthermore, we encourage the readers to go through the supplementary material for few more analysis and possible future works.

%\textbf{Running time comparison or at least for our method}

%=============================
\section{Conclusion}\label{ss:conclusion}
\noindent To reconstruct a dense 3D model of a complex, dynamic, and generally non-rigid scene from its two images captured by an arbitrarily-moving monocular camera is often considered as a very challenging task in Structure-from-Motion. In contrast, the reconstruction of a rigid and stationary scene from two views is a mature and standard task in 3D computer vision, which can be solved easily if not trivially.

This paper has demonstrated that such a dense 3D reconstruction of dynamic scenes is, in fact possible, provided that certain prior assumptions about the scene geometry and about the dynamic deformation of the scene are satisfied. Specifically, we only require that 1) the dynamic scene to be reconstructed is piecewise planar, and 2) the deformation itself between the two frames is locally-rigid but globally as-rigid-as-possible. Both assumptions are mild and realistic, commonly satisfied by real-world scenarios. Our new method dubbed as {\em the SuperpixelSoup algorithm} is able to solve such a challenging problem efficiently, leading to accurate and dense reconstruction of complex dynamic scenes. We hope in theory our method offers a valuable new insight to monocular reconstruction, and in practice, it provides a promising means to perceive a complex dynamic environment by using a single monocular camera. Finally, we want to stress that the rigidity assumption (and the {\small ARAP} constraint) used by the paper is a powerful tool in multi-view geometry research---careful investigation of which may open up new opportunities in the development of advanced techniques for 3D reconstruction.

\section*{Acknowledgment}
\noindent This work was supported in part by Australian Research Council (ARC) grants (DE140100180, DP120103896, LP100100588, CE140100016), Australia ARC Centre of Excellence Program on Robotic Vision, NICTA (Data61) and Natural Science Foundation of China (61420106007). We thank the AC and the reviewers for their invaluable suggestions and comments.

\balance
{\small
\bibliographystyle{ieee}
\bibliography{Dense3D-Reference}
\nocite{kumar2016multi}
\nocite{kumar2017spatio}
\nocite{igarashi2005rigid}
}

\begin{figure*}
\centering
\includegraphics[width=1.0\textwidth, height=0.27\textwidth]{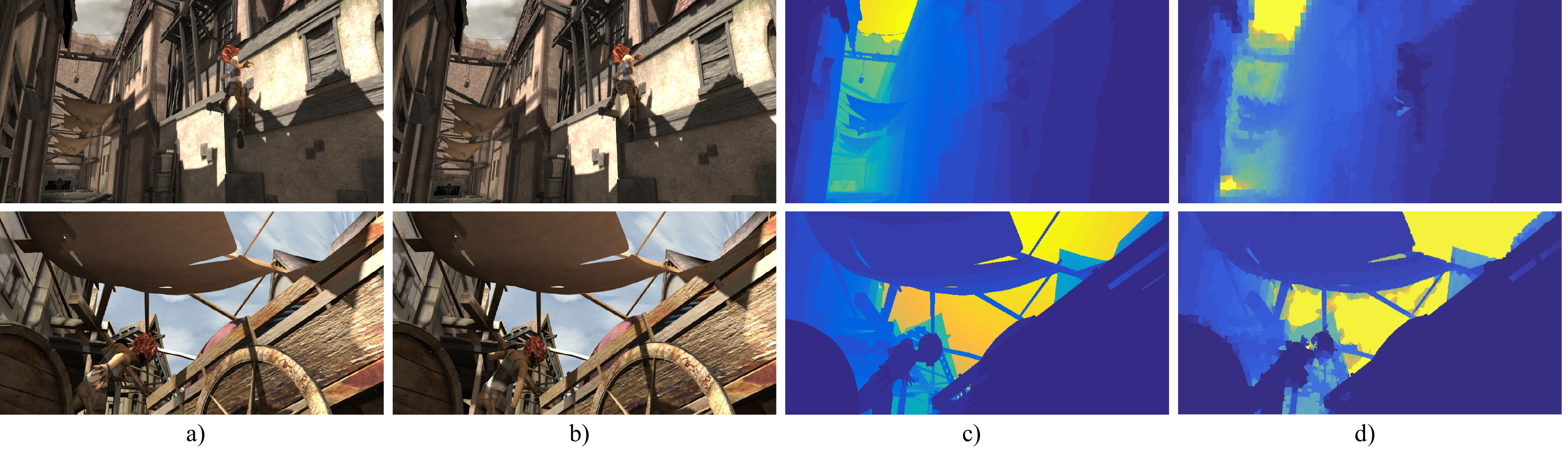}
\caption{\small (a)-(b) are the reference frame and the next frame. It is a very challenging case for proper relative scale recovery with monocular images with dynamic motion. In both cases, the motion of the girl between two consecutive frames is very large and therefore the neighboring relations of planes (say superpixels in image domain) in the consecutive frames get violated. In such cases, our method may not be able to provide correct relative scales for each moving planes in 3D. As a result, the complicated motion of the feet of the girl in this example cannot be explained correctly. In the second example, the cart along with girl is moving w.r.t the camera. The hand of the girl has a substantial motion in consecutive frames. (c)-(d) are the ground-truth and obtained depth map respectively. (Best Viewed on Screen)}
\label{fig:failurecase}
\end{figure*}
\appendix
\balance
\section{Supplementary Material: Monocular Dense 3D Reconstruction of a Complex Dynamic Scene from Two Perspective Frames.}
\subsection{Discussion on failure cases}
The success of our method depends on the effectiveness of the piece-wise planar and rigid motion assumption. Our method may fail if the piece-wise smooth model is no longer a valid approximation for the dynamic scene. 

Furthermore, our approach may also disappoint if the motions of the dynamic objects in the scene between consecutive frames are significantly large such that the neighboring relations defined in the reference frame get violated in the next frame. A couple of examples for such situations are illustrated in Fig.\ref{fig:failurecase}. Other possible situations of failure may arise in the case of textureless surfaces.  Interested readers, researchers, and critics may refer to some new source of information, such as examining surface shading for surface description \cite{blake1985surface}. However, we would like to argue that our algorithm assumes that reasonable dense feature correspondences are provided as input.
\subsection{Future work and possible extension}
One possible extension of our present work is to exploit the current formulation for multiple frames. Other viable and challenging problem is to simultaneously solve for dense optical flow estimation and dynamic 3D reconstruction. A detailed discussion on the aforementioned problems is beyond the scope of this work, however, we want to posit that the solution to these challenging problems is important for the development of sophisticated dense reconstruction algorithm.

\end{document}